\documentclass[times, review, 10pt]{elsarticle}

\usepackage{amssymb}
\usepackage{amsmath}
\usepackage{graphicx}

\usepackage{subfigure}
\usepackage{graphicx}  

\usepackage{graphicx}
\usepackage{booktabs}
\usepackage{wrapfig}
\usepackage{xspace}

\usepackage[marginal]{footmisc}
\usepackage{bm}
\usepackage{textcomp}
\usepackage{gensymb}
\usepackage{pifont}
\usepackage[T1]{fontenc}
\usepackage{multirow}
\usepackage[rightcaption]{sidecap}
\usepackage{makecell}
\usepackage{xspace}
\usepackage{enumitem}
\usepackage{url}
\usepackage{enumerate}
\usepackage{algorithm}
\usepackage[noend]{algpseudocode} 
\usepackage{wrapfig,lipsum,booktabs}
\usepackage{CJKutf8}

\usepackage{hyperref}
\hypersetup{breaklinks,colorlinks}
\usepackage{natbib}

\newcommand{\blue}{}

\usepackage[capitalize]{cleveref}
\crefname{section}{Sec.}{Secs.}
\Crefname{section}{Section}{Sections}
\Crefname{table}{Table}{Tables}
\crefname{table}{Tab.}{Tabs.}

\makeatletter
\DeclareMathOperator*{\argmax}{arg\,max}

\DeclareRobustCommand\onedot{\futurelet\@let@token\@onedot}
\def\@onedot{\ifx\@let@token.\else.\null\fi\xspace}
\def\eg{\emph{e.g}\onedot} 
\def\ie{\emph{i.e}\onedot}

\def\etal{\emph{et al}\onedot}

\makeatother

\journal{Pattern Recognition}

\begin{document}

\begin{frontmatter}



\title{ { Thought Graph Traversal for Test-time Scaling \\ in Chest X-ray VLLMs}}



\author{Yue Yao\textsuperscript{a,1,*}, Zelin Wen\textsuperscript{a,1}, Yan Tong\textsuperscript{a,1}, Xinyu Tian\textsuperscript{b}, \\ Xuqing Li\textsuperscript{a}, Xiao Ma\textsuperscript{a}, Dongliang Xu\textsuperscript{a}, Tom Gedeon\textsuperscript{b,c}}

\affiliation[a]{%
  organization={Shandong University},
  addressline={No. 17923 Jingshi Road},
  city={Jinan},
  postcode={250061},
  state={Shandong},
  country={China}
}

\affiliation[b]{%
  organization={Australian National University},
  addressline={Acton Campus},
  city={Canberra},
  postcode={2601},
  state={ACT},
  country={Australia}
}

\affiliation[c]{%
  organization={Curtin University},
  addressline={Kent Street}, 
  city={Perth},
  postcode={6102}, 
  state={Western Australia},
  country={Australia}
}

\fntext[equal]{\textsuperscript{1}Equal contribution.}
\fntext[equal]{\textsuperscript{*} Corresponding author.}




\begin{abstract}
Test-time scaling offers a promising way to improve the reasoning performance of vision-language large models (VLLMs) without additional training. In this paper, we explore a simple but effective approach for applying test-time scaling to chest X-ray report generation. Specifically, we introduce a lightweight Thought Graph Traversal (TGT) framework that guides the model to reason through organ-specific findings in a medically coherent order. This framework integrates structured medical priors into the prompt, enabling deeper and more logical analysis with no changes to the underlying model. To further enhance reasoning depth, we apply a reasoning budget forcing strategy that adjusts the model's inference depth at test time by dynamically extending its generation process. This simple yet powerful combination allows a frozen radiology VLLM to self-correct and generate more accurate, consistent chest X-ray reports. Our method outperforms baseline prompting approaches on standard benchmarks, and also reveals dataset biases through traceable reasoning paths. Code and prompts are open-sourced for reproducibility at \url{https://github.com/glerium/Thought-Graph-Traversal}

\end{abstract}





\end{frontmatter}



\section{Introduction}

\label{sec:intro}

\begin{figure}[t]
\centering
\includegraphics[width=\columnwidth]{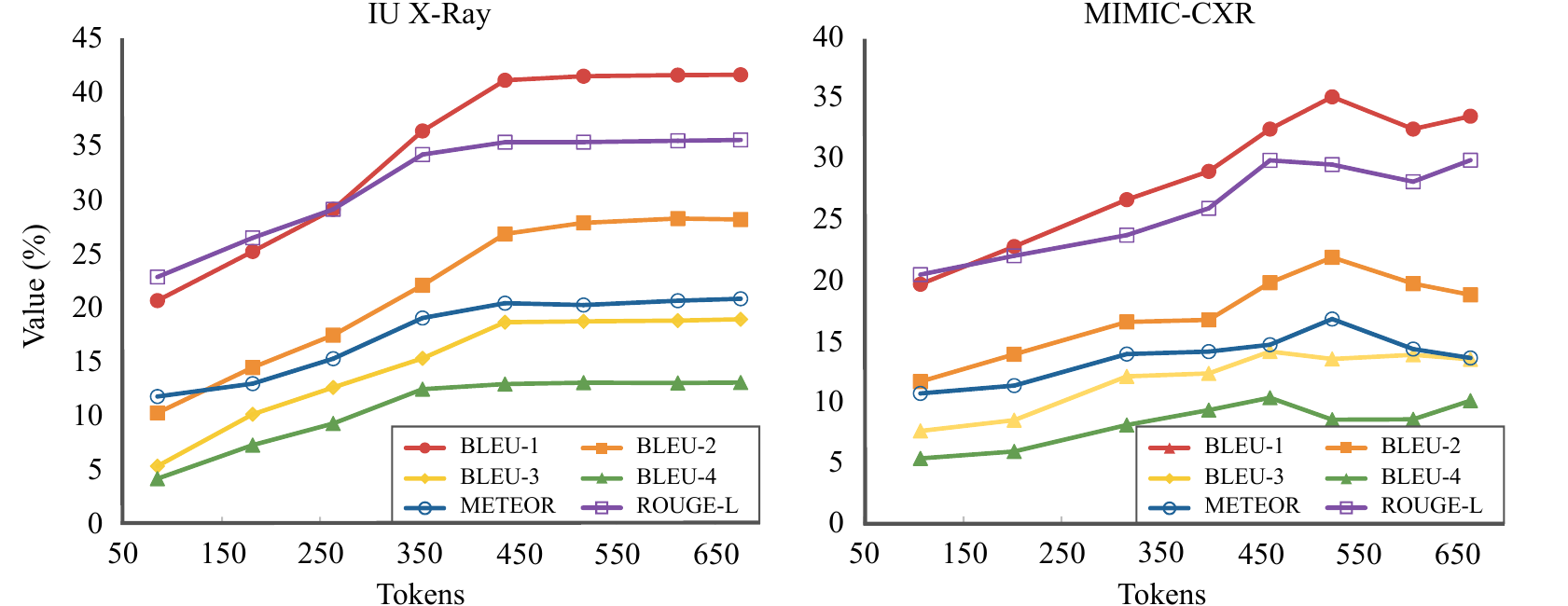}
{\caption{\textbf{Test-time scaling with thought graph traversal.} We apply test-time scaling to chest X-ray report generation by introducing a thought graph traversal framework, which enables structured, multi-step reasoning under varying test-time compute budgets. Our method improves report quality when reasoning about a budget increase (measured by the length of reasoning tokens). Notably, model accuracy shows a positive correlation with the number of reasoning tokens used during inference, until it reaches a saturation point beyond which additional reasoning yields diminishing returns. To balance quality and efficiency, we therefore set the reasoning-token budget near this saturation point during inference.}}
\label{fig:scaling}
\end{figure}

Performance gains in vision-language large models in recent years have predominantly stemmed from increasing training-time computational resources, particularly through the collection of large-scale datasets for self-supervised pretraining~\citep{hoffmann2022trainingcomputeoptimallargelanguage}. These advancements have enabled a shift toward a new paradigm: \textit{test-time scaling}, which focuses on enhancing model performance by increasing computation during inference~\cite{muennighoff2025s1,wang2025ctpt,yin2025context}. A growing body of work like Snell \etal has explored this concept~\citep{snell2024scalingllmtesttimecompute}, and its effectiveness was recently confirmed by OpenAI's o1~\citep{o1}, which achieved notable reasoning improvements by scaling test-time compute. OpenAI attributes this to large-scale reinforcement learning (RL), which implies extensive use of data~\citep{o1}. In response, various groups have attempted to replicate the success of o1 using strategies like Monte Carlo Tree Search~\citep{zhang2024o1codero1replicationcoding}, multi-agent collaboration~\citep{qin2024o1replicationjourneystrategic}, and other techniques like multi-stage reinforcement learning in DeepSeek R1~\citep{r1}. Nevertheless, a task-specific implementation of test-time scaling in a specific area (\ie, medical report generation) has yet to be widely studied. 

Our approach is also motivated by recent research in medical report generation. Many existing methods like Chen~\etal employ encoder-decoder architectures~\citep{chen2020generating}. These methods aim to produce complete radiology reports from input images. However, several critical shortcomings remain unaddressed. \textbf{Structural deficiency:} many generated radiology reports lack standardized structure, reducing readability and making it harder for clinicians to identify essential findings~\citep{ganeshan2018structured}. This issue is exacerbated by training on datasets composed of inconsistent and heterogeneous report formats. As a result, uniformity and clarity in generated outputs are compromised. \textbf{Lack of interpretability and interactivity:} current systems often fail to provide transparent reasoning, relying instead on attention maps that offer limited insights. Furthermore, they lack mechanisms for clinicians to guide or modify the report generation process in accordance with specific patient contexts. This lack of transparency and flexibility limits their usefulness in real-world settings~\citep{miller2019explanation}.

Addressing these limitations, we propose a thought graph traversal (TGT) that explicitly manages the depth of reasoning within VLLMs during inference. By integrating a dedicated reasoning mechanism, our model dynamically controls the extent of logical reasoning at test time, enabling deeper and more clinically relevant analyses. Additionally, thought graph traversal allows structured prior medical knowledge, systematically guiding the model towards logically coherent and medically accurate outputs. We conduct extensive ablation experiments targeting the structured reasoning of VLLMs' results. An additional benefit of this module is its capacity to expose underlying biases present in chest X-ray datasets, which is crucial for improving model generalizability and fairness.

Through extensive experimentation, our method demonstrates superior performance compared to baseline models across established benchmark datasets (\eg, MIMIC-CXR). Shown in Figure~\ref{fig:scaling}, by implementing reasoning budget forcing, we successfully scale the model’s inference capabilities, resulting in significant improvements in diagnostic accuracy of generated reports. To foster transparency, reproducibility, and community-driven advancements, our developed model, along with datasets and source code, are publicly available.


\section{Related work}

\noindent \textbf{Vision large language model.} 
Recent advances in Transformer and computing have enabled very large language models (LLMs) like ChatGPT \cite{openai2023gpt4} with enhanced performance on text generation and translation. Architectures like CLIP \cite{radford2021learning} achieve multimodal understanding via image-text pretraining. Additionally, recent work guides LLMs via prompts rather than explicit training. For example, VisualChatGPT \cite{wu2023visual} connects ChatGPT with vision models for image-inclusive conversations, exemplifying this more flexible prompt-based approach to unleashing LLM potential. The idea of guiding LLMs through prompts has inspired our research. Guiding an LLM through anatomical region prompts results in structured, interpretable reports. Furthermore, clinical context prompts facilitate physician interactivity, enhancing the practical utility of the generated reports.

\noindent \textbf{Radiology VLLMs.} Recent frameworks like Wang~\etal build on medical image captioning to incorporate expert or external knowledge for richer contextual understanding in generated radiology reports \cite{wang2023metransformer}. Recent approaches like the MET \cite{wang2023metransformer}, Kiut \cite{wang2023metransformer}, and KBMA \cite{yang2023radiology} incorporate specialized knowledge through fusion of expert tokens, multimodal injection, and trainable knowledge bases to enhance contextual understanding in radiology report generation. Tanida \etal \cite{tanida2023interactive} uses anatomical regions for the first time to generate reports. To further increase interpretability, some efforts like Qin \etal introduce memory networks \cite{qin2022reinforced} to store knowledge and learn implicit image-text links. Unlike these methods focusing on model training-time strategies, we focus on model test-time scaling only.

\noindent \textbf{Knowledge graphs} have been widely used to incorporate domain knowledge for assisting radiology report generation. For example, Liu \etal proposed integrating pre-constructed graphs that capture relationships between diseases and organs via graph neural networks, enabling dedicated abnormality-aware feature learning \cite{LWG21}. Li \etal later extended this idea by dynamically updating the graph with new knowledge during training \cite{LLC23}. More recently,  Huang \etal designed a knowledge distillation module that fuses information from a symptom graph into the final decoding stage, which shares conceptual similarities with the dynamic decision prompting (DDP) in our work \cite{HZZ23}. However, all these approaches rely on modifying the model through training or finetuning. In contrast, our method performs test-time scaling only, leaving the VLLM architecture and weights untouched. We use a structured prompting mechanism based on a thought graph traversal, which guides the model’s reasoning process in a lightweight and plug-and-play fashion. Despite its simplicity, our method achieves stronger clinical efficacy without requiring additional training or parameter updates.

\noindent \textbf{Prompt engineering and test-time scaling.} Prompt engineering usually integrates human-like problem-solving knowledge~\cite{muennighoff2025s1}. For example, few-shot prompting \cite{brown2020language} allows language models (LMs) to generate responses by providing them with explicit examples. The chain-of-thought (CoT) method \cite{wei2022chain}, along with its variants such as zero-shot CoT \cite{kojima2022large}, graph-of-thought (GoT) \cite{liu2024towards,hu2024prompting}, and tree-of-thought (ToT) \cite{yao2023tree}, intricately designs prompts to emulate various types of human-like reasoning processes. Some other works like Gou \etal \cite{gou2024critic} manually craft prompts that encourage LLMs to engage in critical thinking and verification processes before delivering the final answer. Recent works begin to use prompt engineering for achieving test-time scaling~\cite{muennighoff2025s1}. However, they focus on general methods without a design for a domain-specific area like medical report generation. We, for the first time, design a test-time scaling (\ie, TGT) framework for chest X-ray VLLM. 


\section{Method}

\begin{figure}[t]
\centering
\includegraphics[width=\columnwidth]{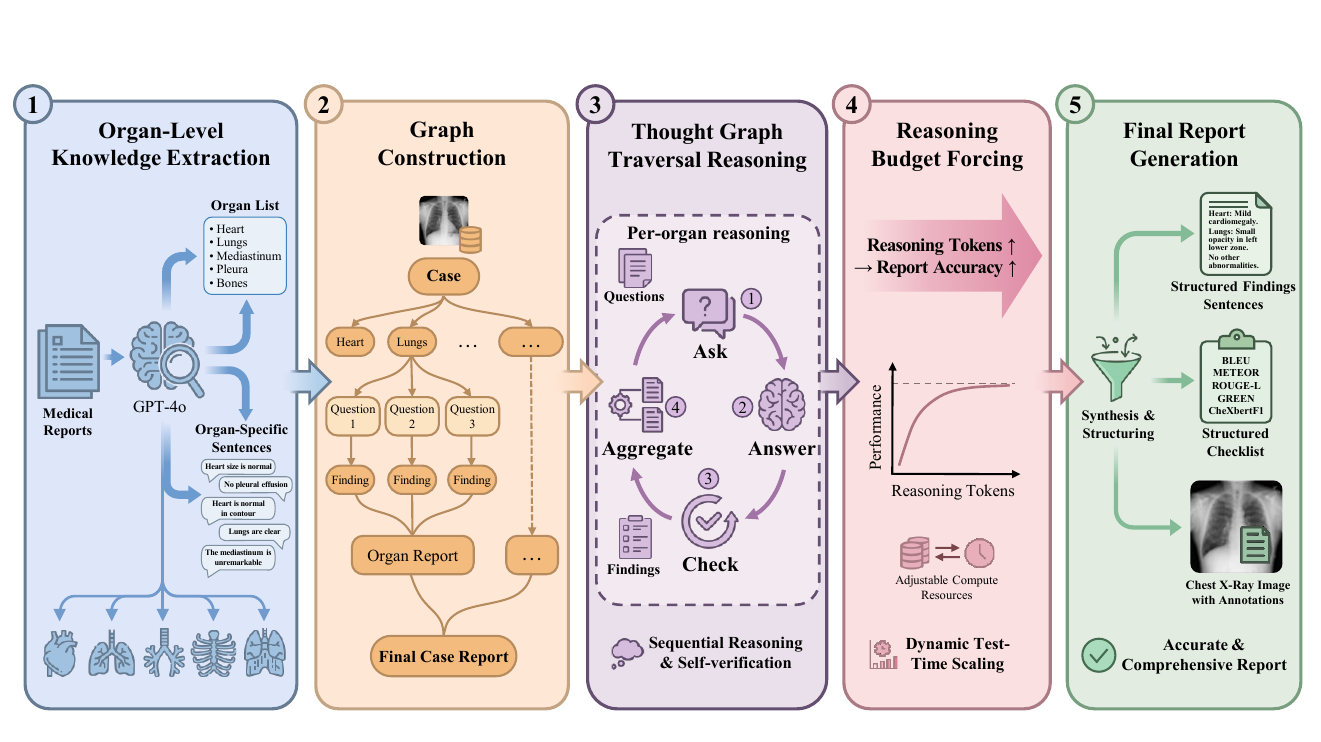}
\vspace{-3em}
\blue{
\caption{
\blue{\textbf{Overview of thought graph traversal for structured chest X-ray report generation.} In the preprocessing \textbf{knowledge extraction} stage, GPT-4o is used to extract organ entities and their corresponding descriptions from training reports, which are stored in the retrieval dictionary. 
In \textbf{graph construction}, for each patient, a fixed set of questions is asked per organ (orange boxes), generating diverse viewpoints on an organ independently. 
These intermediate answers are then aggregated in the \textbf{reasoning} stage, where the VLLM integrates all information to generate an organ-specific statement using in-context examples. Prompt templates are dynamically filled with organ names, retrieved expert examples, and question-answer pairs iteratively with a \textbf{budget forcing} strategy. 
The \textbf{final report} is produced by concatenating the generated statements in a logical order across organs.}
}
}
\label{fig:graph_of_thoughts}
\end{figure}

\subsection{Problem Definition}

We formulate the medical report generation task as a conditional sequence generation problem. Given a chest X-ray image $\mathbf{x}$, the objective is to generate a textual report $\mathbf{y} = (y_1, y_2, \dots, y_T)$, where $y_t$ denotes the $t$-th token in the output sequence. Formally, the prediction is obtained by
\begin{equation}
\hat{\mathbf{y}} = \argmax_{\mathbf{y}} P(\mathbf{y} | \mathbf{x}; \theta),
\end{equation}
where $\theta$ are the model parameters. \blue{Recent advances in VLLMs have demonstrated promising performance in radiology report generation. 
Nevertheless, existing approaches still suffer from several critical limitations, which hinder their effectiveness in clinical settings:
}

\begin{itemize}
\item \textbf{Unstructured output:} Reports $\mathbf{y}$ generated from heterogeneous training sets lack a standardized format, reducing clinical clarity.
\item \textbf{Weak reasoning:} Existing models optimize log-likelihood over $P(\mathbf{y}|\mathbf{x})$, \ie, directly output ground truth only, but lack mechanisms to perform any reasoning.
\end{itemize}

\blue{Our goal is controllable, reasoning-centric generation at test time rather than a fixed end-to-end mapping from images to reports. Therefore, we do not build upon a fully fine-tuned, task-specific report generator and instead construct our framework on top of general-purpose instruction-tuned models. }

\blue{Though existing fully fine-tuned encoder–decoder report generators can produce fluent reports, their high accuracy comes from fixed input and output formats~\cite{10.1007/978-3-032-04978-0_58,Liu_2025,wu2025diseaseawaredualstageframeworkchest}. Thus, the model often succeeds by pattern completion rather than explicit, controllable reasoning. Specifically, these models usually adopt encoder-decoder architectures to generate reports from medical images and are typically fine-tuned on narrow domain-specific corpora, which limits their flexibility in follow complex textual prompts. Their decoding process is often rigid and solely optimized for next-token prediction, which limits their capacity for test-time control or explicit reasoning. Moreover, due to the lack of native instruction-following capabilities and insufficient pretraining on general-purpose reasoning tasks, prompt engineering has limited effect on steering their output structure or depth of inference.}

\blue{Therefore, we do not rely on such architectures and instead build our framework on top of general-purpose, instruction-tuned large language models (\eg, Qwen series~\cite{bai2025qwen25vltechnicalreport} and HuatuoGPT-Vision~\cite{chen2024huatuogptvisioninjectingmedicalvisual}) with strong multimodal alignment and reasoning ability, allowing for controllable, multi-step report generation at test time.}



In this paper, we propose an approach that explicitly manages the depth of reasoning within general-purpose VLLMs during inference, as depicted in Figure~\ref{fig:graph_of_thoughts}. We conduct extensive ablation studies targeting the structured reasoning of VLLM's results and test-time scaling. By integrating a dedicated reasoning mechanism, our model dynamically controls the extent of logical reasoning at test time, enabling deeper and more clinically relevant analyses. To this end, we introduce a thought graph traversal framework enriched with structured prior medical knowledge, systematically guiding the model towards logically coherent and medically accurate outputs. An additional benefit of this module is its capacity to expose underlying biases present in chest X-ray datasets, which is crucial for improving model generalizability and fairness.

Specifically, we introduce: (1) A thought graph traversal module that infuses structured medical knowledge $\mathcal{K}$ to guide report generation. (2) A reasoning budget control mechanism, allowing dynamic adjustment of reasoning depth. With these components, our approach moves beyond surface-level generation and enables deeper, more interpretable, and more consistent diagnostic outputs.

\begin{figure}[H]
\centering
\includegraphics[width=\columnwidth]{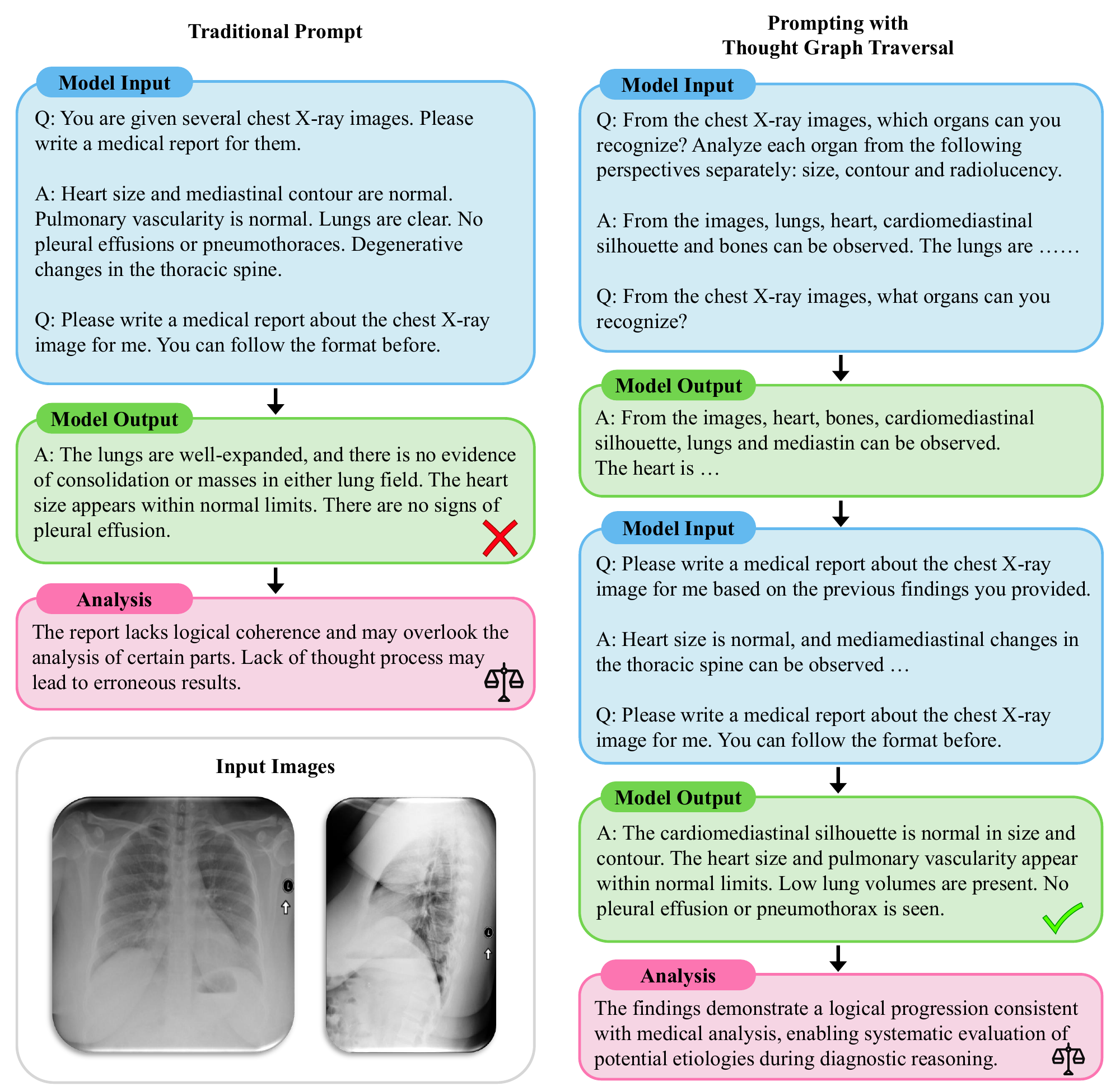}
\vspace{-2em}
\blue{\caption{\textbf{Examples of conventional prompting and our thought graph traversal.} \textbf{Left:} Conventional prompting mimics stylistic examples without deep understanding, often leading to hallucinations or logically inconsistent reports. \textbf{Right:} Our method explicitly guides the model’s attention toward organ-level reasoning and enforces a medically coherent reasoning order, leading to more accurate, logical, and interconnected diagnostic statements in the generated report. }}
\label{fig:prompt_compare}
\end{figure}


\subsection{Test-Time Reasoning via Graph Traversal}
\label{sec:graph}

We classify our graph traversal into \textbf{1) sequential traversal}, where later computations depend on earlier ones (\ie, a long reasoning trace), and \textbf{2) parallel traversal}, where computations run independently (\ie, majority voting). We choose sequential scaling first as, intuitively, this approach is expected to scale better, since later computations can build on intermediate results, allowing for deeper reasoning and iterative refinement. We propose new sequential scaling methods and ways to benchmark them. Our main process for report generation is shown in Algorithm \ref{alg:report-generation}. Our data consists of multiple X-ray images per patient as input $\{\mathbf{x}_1, \dots, \mathbf{x}_n\}$ and a corresponding report $\mathbf{y}$, which includes several sentences, each describing a particular organ. The goal is to generate $\mathbf{y}$ from the image set $\{\mathbf{x}_i\}$ using a vision-language model.

\begin{algorithm}[t]
\caption{Thought Graph Traversal}
\label{alg:report-generation}
\begin{algorithmic}[1]
\Require Patient X-ray images $\{\mathbf{x}_i\}$, an organ-indexed sentence retrieval dictionary $\mathcal{D}$, with organ set $\mathcal{O}$
\Ensure Full medical report $\mathbf{y}$
\State Initialize $\mathbf{y} \gets \emptyset$
\For{each organ $o \in \mathcal{O}$}
    \State $Q \gets \texttt{GetQuestions}(o)$ \Comment{Retrieve diagnostic questions}
    \State $\mathcal{C}_o \gets \emptyset$
    \For{each question $q \in Q$}
        \State $Verified \gets \texttt{False}$
        \While{not $Verified$}
            \State $a \gets \texttt{VLLM.Answer}\left(\{\mathbf{x}_i\}, q\right)$
            \State $Verified \gets \texttt{VLLM.Verify}\left(\{\mathbf{x}_i\}, q, a\right)$ \Comment{Verification Step}
            \EndWhile
    \EndFor
        \State $\mathcal{E}_o \gets \texttt{SampleExamples}\left(\mathcal{D}_o, k=5\right)$ \Comment{Few-shot ICL examples}
        \State $\mathcal{C}_o \gets \mathcal{C}_o \cup \{(q,a)\}$
        \State $r_o \gets \texttt{VLLM.GenerateReport}\left(\{\mathbf{x}_i\}, \mathcal{C}_o, \mathcal{E}_o\right)$

    \State Append $r_o$ to $\mathbf{y}$
\EndFor
\State \Return $\mathbf{y}$
\end{algorithmic}
\end{algorithm}



\blue{Before inference, we perform entity extraction to construct an organ-indexed retrieval dictionary $\mathcal{D}
= \{ \mathcal{D}_o \mid o \in \mathcal{O} \}$, where $\mathcal{D}_o$ denotes the set of sentences associated with organ $o$. We divide the entire dataset into a pseudo-training set and a pseudo-testing set. Notably, no VLLM training is involved in this stage; instead, the pseudo-training set is used to extract prior knowledge for structured reasoning. Specifically, we employ GPT-4o to extract all organ entities mentioned in training reports, forming an ordered organ set $\mathcal{O} = \{o_1, o_2, \dots, o_n\}$, which serves as the key set of the retrieval dictionary. For each organ $o \in \mathcal{O}$, we collect its associated textual descriptions from the training reports. These organ-wise sentences serve as in-context learning (ICL)~\cite{dong-etal-2024-survey} examples for later inference.}


We formulate the structured report generation process as a reasoning traversal over a conceptual graph $\mathcal{G}$, where each node corresponds to an organ $o \in \mathcal{O}$, and each node branches into \blue{several} child nodes representing domain-specific diagnostic questions \blue{$q \in Q$} \blue{where each question $q$ is constructed by performing an in-depth analysis of medical reports to select the most relevant diagnostic aspects for that node $o$ from the pre-processed retrieval dictionary. This process ensures that each question accurately corresponds to the key observations and diagnostic elements in a clinical report.}.

\textbf{Graph Construction.} The graph $\mathcal{G}$ is constructed as a two-level hierarchy. The root node corresponds to the entire patient case. The first level consists of organ nodes \blue{$\mathcal{O}$} extracted from training data in the entity extraction process. Each organ node $o$ connects to a fixed sequence of diagnostic question nodes \blue{$Q$}, forming a reasoning subgraph for that organ.

\textbf{Traversal Strategy.} \blue{During inference, we adopt a two-level traversal of the graph. Starting from the root, we iterate through all organ nodes $o \in \mathcal{O}$ (\ie, in parallel over organs). For each organ, its associated question nodes $q \in Q$ are expanded, and each question is paired with the patient images $\{\mathbf{x}_i\}$ to prompt the VLLM for an answer. After an initial answer $a$ is generated, a verification step is performed to ensure its consistency with the question $q$ and the input images $\{\mathbf{x}_i\}$ \blue{by asking the VLLM whether $a$ correctly answers $q$ and accurately describes the content in images $\{\mathbf{x}_i\}$}. Unverified answers are regenerated until the verification succeeds. All verified question–answer pairs accumulated during traversal are collected into a reasoning context $\mathcal{C}_o = \{(q, a)\}$ for organ $o$. Finally, conditioned on the images $\{\mathbf{x}_i\}$, the reasoning context $\mathcal{C}_o$, and retrieved in-context examples $\mathcal{E}_o$, the VLLM produces the organ-level report $r_o$.
}


\textbf{Report Assembly.} The final report $\mathbf{y}^*$ is obtained by sequentially concatenating the generated descriptions $r_o$ in the original organ order. $\mathbf{y}^* = [r_{o_1}, r_{o_2}, \dots, r_{o_n}].$ This structured, graph-aware design provides several key benefits. By making each organ’s reasoning module self-contained, the framework scales naturally: organ-level reasoning can be executed in parallel and extended to additional organs or question types with minimal changes. At the same time, the BFS-style traversal aligns with common clinical reading patterns, yielding a transparent reasoning path that exposes intermediate decisions and supports step-by-step inspection. Finally, the procedure is controllable at the prompt level, enabling users to adjust the depth of questioning, skip irrelevant organs, and prune low-confidence branches to trade off coverage, reliability, and computation.

\subsection{Budget Forcing via Iterative Traversal}

Building on the graph-based reasoning framework in Section~\ref{sec:graph}, we further introduce a budget forcing mechanism to regulate the amount of computation during test-time traversal. While Section~\ref{sec:graph} adopts a parallel-sequence framework over the structured reasoning graph $\mathcal{G}$ to cover all diagnostic questions per organ, here we refine that process by limiting how deep or how far the traversal proceeds, \ie, how many reasoning tokens the model is allowed to consume. Specifically, as described in the last section, instead of generating all question-answer pairs in a single pass, we initiate a multi-round inference process. In each round, follow-up questions are asked based on prior responses, aiming to explore new diagnostic angles and progressively deepen reasoning. This iterative reasoning is still rooted in the BFS-style traversal introduced earlier but adds a temporal dimension: each organ node now accumulates its subgraph traversal over time, guided by token budget constraints.

As illustrated in Figure~\ref{fig:scaling}, we observe a clear positive correlation between the number of reasoning tokens and report quality. However, model performance plateaus around 450 tokens, indicating that additional reasoning beyond this point yields minimal improvement. This saturation point provides a practical guideline for balancing inference cost and quality under test-time compute constraints.

To ensure the reliability of the generated content, we further introduce a final verification step. After producing an initial sentence $r_o$ for each organ, the VLLM revisits the reasoning path taken (\ie, $\mathcal{C}_o$) and evaluates whether the generated output is coherent and justified. If inconsistencies are detected—such as unsupported conclusions or logical contradictions—the system re-enters the reasoning loop for that organ, repeating the traversal with updated prompts until a satisfactory report is obtained.






\begin{table}
\centering
\scriptsize
\setlength{\tabcolsep}{0.7mm}
\caption{\textbf{Comparison of Zero-shot, traditional Few-shot, and thought graph traversal on the IU X-Ray dataset.} The table reports evaluation metrics including BLEU-1 to BLEU-4, METEOR, ROUGE-L, \blue{GREEN and CheXbertF1} across three models (GPT-4o \cite{openai2024gpt4ocard}, Qwen2.5-VL\blue{-7B} \cite{bai2025qwen25vltechnicalreport}, and HuatuoGPT-Vision\blue{-7B} \cite{chen2024huatuogptvisioninjectingmedicalvisual}). Thought Graph Traversal (TGT) consistently achieves higher scores, indicating improved report generation quality.}

\begingroup
\renewcommand{\arraystretch}{1.2}
\begin{tabular}{c|c|cccccccc} 
\Xhline{1.2pt}
\multicolumn{1}{l}{}       &                     & \textbf{BLEU-1}  & \textbf{BLEU-2}  & \textbf{BLEU-3}  & \textbf{BLEU-4}  & \textbf{METEOR}  & \textbf{ROUGE-L} & \blue{\textbf{GREEN}} & \blue{\textbf{CheXbertF1}} \\ 
\hline
\multirow{3}{*}{\rotatebox{90}{GPT-4o}}       & Zero-shot       & \blue{20.55$\pm$0.30}    & \blue{9.87$\pm$0.28}     & \blue{5.36$\pm$0.14}    & \blue{2.58$\pm$0.23}     & \blue{15.86$\pm$0.20}   & \blue{20.59$\pm$0.19}    &   \blue{65.97$\pm$0.37}   & \blue{71.24$\pm$1.38}\\
& Few-shot     & \blue{35.74$\pm$0.28}    & \blue{22.42$\pm$0.19}    & \blue{14.85$\pm$0.13}   & \blue{9.44$\pm$0.15}     & \blue{20.29$\pm$0.17}   & \blue{28.71$\pm$0.16}   & \blue{75.15$\pm$0.28}    & \blue{77.82$\pm$0.65}       \\
& TGT         & \blue{\textbf{40.45}$\pm$0.45}    & \blue{\textbf{27.48}$\pm$0.34}     & \blue{\textbf{17.55}$\pm$0.26} & \blue{\textbf{11.81}$\pm$0.25}     & \blue{\textbf{20.24}$\pm$0.17}   & \blue{\textbf{35.17}$\pm$0.21}     & \blue{\textbf{80.02}$\pm$0.29}   & \blue{\textbf{78.24}$\pm$0.43 }     \\
\hline

\multirow{3}{*}{\rotatebox{90}{Qwen2.5-VL}}      & Zero-shot         & \blue{19.13$\pm$0.63}    & \blue{8.70$\pm$0.30}     & \blue{4.44$\pm$0.22}    & \blue{2.22$\pm$0.20}     & \blue{14.95$\pm$0.13}   & \blue{19.09$\pm$0.33}    & 
\blue{48.54$\pm$3.49}   & \blue{28.02$\pm$4.58} \\
& Few-shot  & \blue{33.37$\pm$0.22}    & \blue{18.97$\pm$0.16}    & \blue{12.06$\pm$0.14}    & \blue{7.61$\pm$0.07}    & \blue{18.19$\pm$0.09}   & \blue{26.67$\pm$0.48}    & \blue{74.02$\pm$0.12}    & \blue{75.03$\pm$0.20}     \\
& TGT       & \blue{\textbf{38.23}$\pm$0.72}    & \blue{\textbf{25.53}$\pm$0.84}    & \blue{\textbf{18.65}$\pm$0.85}    & \blue{\textbf{12.66}$\pm$0.75}    & \blue{\textbf{19.33}$\pm$0.27}    & \blue{\textbf{36.34}$\pm$0.85}    &     \blue{\textbf{75.70}$\pm$0.14}    & \blue{\textbf{77.68}$\pm$1.15} \\ 
\hline

\multirow{3}{*}{\rotatebox{90}{\shortstack{HuatuoGPT-\\Vision}}} & Zero-shot & \blue{18.60$\pm$0.14}    & \blue{8.44$\pm$0.12}     & \blue{4.20$\pm$0.07} & \blue{2.10$\pm$0.03}     & \blue{14.74$\pm$0.11}   & \blue{17.80$\pm$0.22}     & \blue{72.92$\pm$0.59}   & \blue{49.10$\pm$1.67}    \\

& Few-shot  & \blue{33.88$\pm$0.13}    & \blue{24.02$\pm$0.20}     & \blue{12.11$\pm$0.17} & \blue{9.88$\pm$0.12}     & \blue{20.15$\pm$0.14}   & \blue{25.07$\pm$0.09}     & \blue{70.99$\pm$0.27}   & \blue{76.21$\pm$0.10}     \\

& TGT       & \blue{\textbf{43.89}$\pm$0.29}    & \blue{\textbf{28.75}$\pm$0.18}     & \blue{\textbf{19.84}$\pm$0.15} & \blue{\textbf{13.58}$\pm$0.09}     & \blue{\textbf{21.39}$\pm$0.17}   & \blue{\textbf{37.62}$\pm$0.20}     & \blue{\textbf{73.03}$\pm$0.38}   & \blue{\textbf{78.45}$\pm$0.99}     \\
\Xhline{1.2pt}
\end{tabular}
\endgroup

\label{tab:comparison_iuxray}
\end{table}

\section{Experiments}
\subsection{Implementation Details}

IU X-Ray~\cite{pavlopoulos2019survey} is a widely used publicly available dataset for medical report generation tasks containing 3,955 fully de-identified chest X-ray reports with findings, each associated with frontal and/or lateral chest X-rays, totaling 7,470 images.

MIMIC-CXR \cite{johnson2019mimic} is currently the largest public dataset containing many chest X-ray images and reports. In total, this dataset has 377,110 images and 227,835 reports for 64,588 patients. For experimental and fair comparisons, we followed the previous methodology \cite{qin2022reinforced} used the official MIMIC-CXR divisions: 222,758 samples for training, 1,808 for validation, and 3,269 for testing. 

We evaluated X-ray report generation using standard natural language generation (NLG) metrics. The NLG metrics were BLEU \cite{papineni2002bleu}, METEOR \cite{banerjee2005meteor}, and ROUGE \cite{lin2004rouge} scores, which are standard metrics used to assess the fluency of generated natural language. \blue{In addition, we included two clinically oriented evaluation metrics, GREEN \cite{ostmeier2024green} and CheXbertF1 \cite{irvin2019chexpert}, to better measure radiology-specific factual consistency and diagnostic correctness.}


\begin{table}
\centering
\scriptsize
\setlength{\tabcolsep}{0.7mm}
\caption{\textbf{Comparison of Zero-shot, traditional Few-shot, and thought graph traversal prompting on the MIMIC-CXR dataset.} The table reports evaluation metrics including BLEU-1 to BLEU-4, METEOR, ROUGE-L\blue{, GREEN and CheXbertF1} across three models (GPT-4o, Qwen2.5-VL\blue{-7B}, and HuatuoGPT-Vision\blue{-7B}). Values are presented in percentage (\%). Thought graph traversal consistently achieves higher scores, indicating improved report generation quality.}
\begingroup
\renewcommand{\arraystretch}{1.2}  

\begin{tabular}{c|c|cccccccc} 
\Xhline{1.2pt}
\multicolumn{1}{l}{} &                     & \textbf{BLEU-1}  & \textbf{BLEU-2}  & \textbf{BLEU-3}  & \textbf{BLEU-4}  & \textbf{METEOR}  & \textbf{ROUGE-L} & \textbf{\blue{GREEN}} & \textbf{\blue{CheXbertF1}} \\ 
\hline

\multirow{3}{*}{\rotatebox{90}{GPT-4o}}      
& Zero-shot & \blue{19.53$\pm$0.35} & \blue{7.89$\pm$0.30} & \blue{1.42$\pm$0.22} & \blue{1.05$\pm$0.28} & \blue{9.54$\pm$0.24} & \blue{16.13$\pm$0.26} & \blue{20.24$\pm$0.45} & \blue{21.05$\pm$1.20} \\
& Few-shot  & \blue{24.93$\pm$0.25} & \blue{11.26$\pm$0.18} & \blue{4.39$\pm$0.14} & \blue{3.00$\pm$0.16} & \blue{12.24$\pm$0.18} & \blue{18.09$\pm$0.17} & \blue{36.98$\pm$0.30} & \blue{21.05$\pm$0.70} \\
& TGT       & \blue{\textbf{26.83}$\pm$0.40} & \blue{\textbf{14.37}$\pm$0.32} & \blue{\textbf{8.94}$\pm$0.26} & \blue{\textbf{5.10}$\pm$0.27} & \blue{\textbf{14.50}$\pm$0.18} & \blue{\textbf{21.51}$\pm$0.24} & \blue{\textbf{39.69}$\pm$0.36} & \blue{\textbf{42.11}$\pm$0.50} \\
\hline

\multirow{3}{*}{\rotatebox{90}{Qwen2.5-VL}}
& Zero-shot & \blue{21.06$\pm$0.80} & \blue{8.06$\pm$0.65} & \blue{1.62$\pm$0.55} & \blue{1.10$\pm$0.50} & \blue{10.28$\pm$0.30} & \blue{15.17$\pm$0.40} & \blue{28.75$\pm$3.20} & \blue{10.53$\pm$4.00} \\
& Few-shot  & \blue{22.95$\pm$0.30} & \blue{10.10$\pm$0.25} & \blue{3.43$\pm$0.20} & \blue{1.77$\pm$0.15} & \blue{11.96$\pm$0.18} & \blue{15.99$\pm$0.55} & \blue{36.28$\pm$0.40} & \blue{21.05$\pm$0.50} \\
& TGT       & \blue{\textbf{26.20}$\pm$0.70} & \blue{\textbf{14.86}$\pm$0.85} & \blue{\textbf{8.09}$\pm$0.90} & \blue{\textbf{5.06}$\pm$0.80} & \blue{\textbf{13.99}$\pm$0.35} & \blue{\textbf{21.11}$\pm$0.90} & \blue{\textbf{38.94}$\pm$0.70} & \blue{\textbf{31.58}$\pm$1.20} \\
\hline

\multirow{3}{*}{\rotatebox{90}{\shortstack{HuatuoGPT-\\Vision}}}
& Zero-shot & \blue{22.97$\pm$0.18} & \blue{7.51$\pm$0.14} & \blue{2.52$\pm$0.10} & \blue{1.24$\pm$0.06} & \blue{10.47$\pm$0.12} & \blue{15.64$\pm$0.25} & \blue{29.31$\pm$0.65} & \blue{26.32$\pm$1.80} \\
& Few-shot  & \blue{27.81$\pm$0.14} & \blue{14.81$\pm$0.22} & \blue{6.73$\pm$0.20} & \blue{3.37$\pm$0.14} & \blue{13.89$\pm$0.16} & \blue{18.29$\pm$0.15} & \blue{\textbf{41.70}$\pm$0.40} & \blue{31.58$\pm$0.40} \\
& TGT       & \blue{\textbf{31.46}$\pm$0.30} & \blue{\textbf{17.42}$\pm$0.18} & \blue{\textbf{9.32}$\pm$0.16} & \blue{\textbf{3.56}$\pm$0.10} & \blue{\textbf{14.88}$\pm$0.22} & \blue{\textbf{21.57}$\pm$0.28} & \blue{40.90$\pm$0.55} & \blue{\textbf{36.84}$\pm$1.00} \\
\Xhline{1.2pt}
\end{tabular}

\endgroup
\label{tab:comparison_mimic}
\end{table}

\blue{
Our experiments were conducted using 8 NVIDIA GeForce RTX 3090 GPUs and 64x AMD EPYC 7343 16-Core Processors. 
To evaluate the performance of thought graph traversal, we leveraged GPT-4o~\cite{openai2024gpt4ocard}, Qwen-2.5-VL-7B~\cite{bai2025qwen25vltechnicalreport}, and HuatuoGPT-VL-7B~\cite{chen2024huatuogptvisioninjectingmedicalvisual} as backbone models, with the MIMIC-CXR and IU X-ray datasets.}

\subsection{Results}


{\textbf{The reliability of entity extraction}. To evaluate the reliability of entity extraction, we compared GPT-4o with a rule-based baseline using regular expression-based search on 2,000 reports. The results demonstrate high consistency between the two methods: GPT-4o successfully extracted 1,520 heart-related descriptions, while the regular expression search matched 1,446. Notably, an analysis of the 80 instances identified exclusively by GPT-4o revealed that these were not extraction errors. Instead, they represented complex linguistic variants, such as nuanced anatomical phrasing, that the rigid regular expression failed to capture. This comparison validates that GPT-4o is not only stable but also provides superior linguistic flexibility, ensuring that the impact of any residual extraction inconsistencies on the final results is negligible.}

\textbf{Thought graph traversal enhances reasoning depth and report quality up to a threshold.} We explore the effect of structured question prompting on reasoning depth. By controlling the number of diagnostic questions asked prior to report generation, we indirectly modulate the model’s token budget for reasoning. As Figure~\ref{fig:scaling} demonstrates, increasing the number of targeted diagnostic queries leads to higher report quality up to a threshold (4 queries), beyond which marginal gains plateau. This suggests that early reasoning steps already address core findings, and later queries add limited value due to informational redundancy. 



\begin{table}
\centering
\scriptsize
\setlength{\tabcolsep}{1.0mm}
\caption{\textbf{Comparison of different prompting or scaling strategies, including \textbf{RaR} \cite{Deng2023RephraseAR}, \textbf{On-MP} \cite{deWynter2023OnM}, and \textbf{Self-Refine} \cite{NEURIPS2023_91edff07}, for HuatuoGPT\blue{-Vision-7B} on the IU X-Ray dataset.} Metrics include BLEU-1 to BLEU-4, METEOR, ROUGE-L\blue{, GREEN and CheXbertF1}. Thought graph traversal achieves the best overall performance. Values are presented in percentage (\%).}

\begingroup
\renewcommand{\arraystretch}{1.2}  

\begin{tabular}{c|cccccccc} 
\Xhline{1.2pt}
\textbf{Method} & \textbf{BLEU-1} & \textbf{BLEU-2} & \textbf{BLEU-3} & \textbf{BLEU-4} & \textbf{METEOR} & \textbf{ROUGE-L} & \textbf{\blue{GREEN}} & \textbf{\blue{CheXbertF1}} \\ 
\hline
Self-Refine & \blue{15.33$\pm$0.21}   & \blue{6.88$\pm$0.13}    & \blue{3.49$\pm$0.10}    & \blue{1.75$\pm$0.06}    & \blue{14.31$\pm$0.09}   & \blue{15.94$\pm$0.15}   & \blue{68.89$\pm$0.23} & \blue{48.08$\pm$0.76} \\
On-MP       & \blue{16.84$\pm$0.10}   & \blue{7.03$\pm$0.08}    & \blue{3.52$\pm$0.11}    & \blue{1.75$\pm$0.09}    & \blue{12.55$\pm$0.06}   & \blue{16.51$\pm$0.26}   & \blue{64.55$\pm$0.28} & \blue{54.63$\pm$1.96} \\
RaR         & \blue{31.00$\pm$0.57}   & \blue{14.50$\pm$0.30}   & \blue{7.91$\pm$0.21}    & \blue{4.15$\pm$0.12}    & \blue{14.12$\pm$0.06}   & \blue{22.80$\pm$0.19}   & \blue{58.91$\pm$0.36} & \blue{74.52$\pm$1.29} \\
TGT         & \blue{\textbf{43.89}$\pm$0.29}   & \blue{\textbf{28.75}$\pm$0.18}  & \blue{\textbf{19.84}$\pm$0.15}  & \blue{\textbf{13.58}$\pm$0.09}  & \blue{\textbf{18.39}$\pm$0.17}  & \blue{\textbf{37.62}$\pm$0.20}  & \blue{\textbf{73.03}$\pm$0.38}  & \blue{\textbf{78.45}$\pm$0.99}   \\
\Xhline{1.2pt}
\end{tabular}
\endgroup

\label{tab:comparison_benchmark}
\end{table}

\begin{figure}[t]
\centering
\includegraphics[width=0.96\columnwidth]{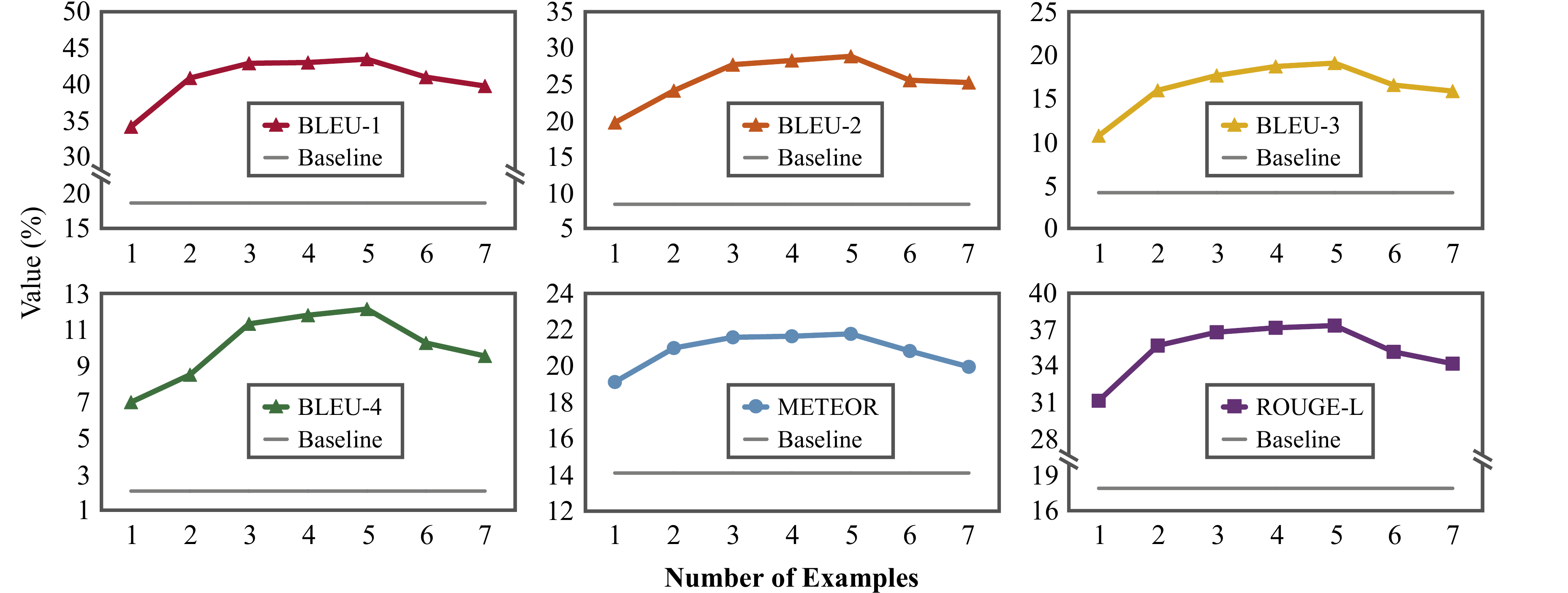}
\blue{\caption{\textbf{Impact of example quantity in prompt design.} We empirically study the effect of example count on model performance during chest X-ray report generation. Using our curated prompt with seven examples as a baseline, we observe that reducing the number of examples significantly degrades report quality due to insufficient guidance, leading to reasoning and formatting errors. Interestingly, increasing the number of examples beyond five also causes a performance drop. This is likely due to the model overfitting to the examples, prioritizing mimicking their patterns over analyzing the actual image, or being misled by repeated or similar examples, thereby reducing factual accuracy.}}
\label{fig:icl_ablation}
\end{figure}

\textbf{Compared to no prompting, our thought graph traversal method introduces structured reasoning that significantly improves report completeness and medical accuracy.} Shown in Table~\ref{tab:comparison_iuxray}, Table~\ref{tab:comparison_mimic}, without any prompt guidance, vision-language models often rely solely on their pretrained biases, leading to generic, underspecified, or irrelevant reports. Thought graph traversal provides a scaffolded inference process by decomposing the task into organ-wise diagnostic reasoning steps, prompting the model to attend to clinically meaningful visual patterns. For example, on the MIMIC-CXR dataset, GPT-4o’s BLEU-4 score improves from \textbf{1.03 (Zero-shot)} to \textbf{9.84 (TTS)}; similarly, HuatuoGPT-Vision improves from \textbf{1.24 to 11.34}, showing substantial gains in output specificity and image-grounding (Table~\ref{tab:comparison_mimic}).

\textbf{Compared to conventional existing prompt engineering methods, thought graph traversal further enhances factual accuracy and logical consistency.} As illustrated in Figure~\ref{fig:prompt_compare}, traditional few-shot prompting, focused on mimicking the format and wording of examples, often causes imitation bias, leading to hallucinated findings and anatomically implausible statements. In contrast, thought graph traversal explicitly guides the model in analyzing the image organ by organ, with each reasoning step informed by representative examples and structured medical knowledge. For instance, HuatuoGPT with thought graph traversal achieves a BLEU-4 score of \textbf{12.14}, outperforming \textbf{RaR (3.25)}, \textbf{On-MP (2.25)}, and \textbf{Self-Refine (2.28)} by large margins across all evaluation metrics (Table~\ref{tab:comparison_benchmark}).

\begin{figure}[t]
\centering
\includegraphics[width=0.96\columnwidth]{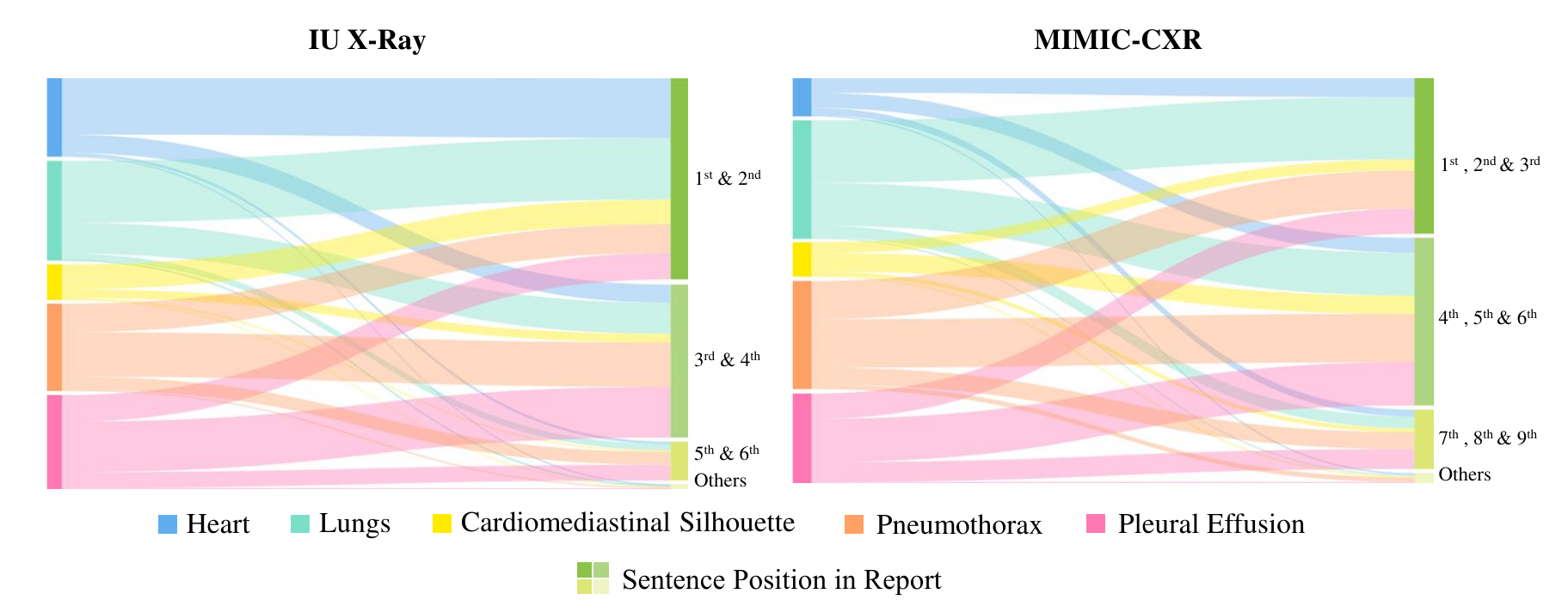}
\caption{\textbf{Analysis of organ description positions in expert-written reports from the IU X-Ray and MIMIC-CXR datasets.} The left y-axis represents the different organs, and the right y-axis indicates the sentence number in the report where each organ is mentioned. The figure shows that certain organs, such as the heart and lungs, tend to appear earlier in reports, while pathological findings are more commonly placed later. This positional pattern suggests that organ-level descriptions follow a latent structural order. To quantify this, we prompt GPT-4o to extract organ-related sentences from each report and pair them with their corresponding organs. These organ-sentence pairs are stored in the retrieval dictionary, enabling visualization and modeling of organ-specific sentence distribution.}
\label{fig:dataset_bias}
\end{figure}

\textbf{Increasing ICL examples improves performance up to a point.} To understand the role of in-context learning (ICL) scale, we vary the number of examples in the prompt. As shown in Figure~\ref{fig:icl_ablation}, performance improves when increasing the number of examples from 1 to 5, but begins to decline beyond 5. We attribute this to cognitive overload: when too many examples are included, the model's attention shifts from visual grounding to excessive example matching, causing reduced clinical validity and misprioritized findings.

\begin{figure}[t]
\centering
\includegraphics[width=0.6\columnwidth]{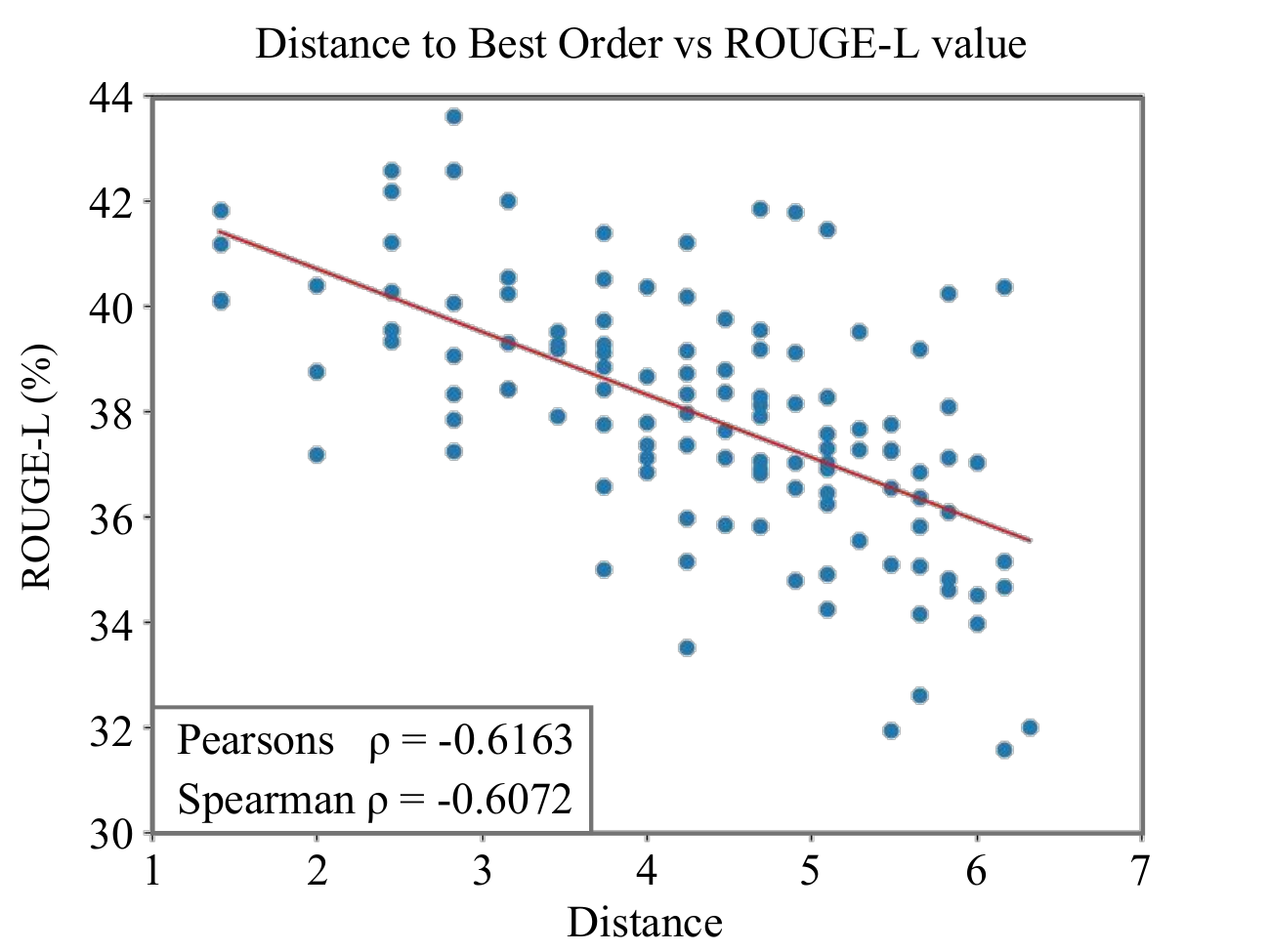}
\caption{\textbf{The scatter plot showing the relationship between organ order distance and the ROUGE-L values.} We conduct an experiment across all \blue{$n!$} permutations of \blue{n} organs in the reasoning graph. Correlation analysis reveals that ROUGE-L is sensitive to organ sequence, with statistically significant negative correlations (Spearman $\rho=-0.6163$, $p=8.53 \times 10^{-14}$; Pearson $\rho=-0.6072$, $p=2.45 \times 10^{-13}$), suggesting the importance of logical ordering in report generation. }
\label{fig:order_metrics_correlation}
\end{figure}

\textbf{Dataset bias on organ description order.} Moreover, our analysis of expert reports reveals a consistent positional bias in organ descriptions. For example, heart and lung findings are frequently mentioned at the beginning of a report, while soft tissue or osseous findings appear later. To leverage this, we use GPT-4o to extract organ-mention pairs from reports, organizing them into the retrieval dictionary. This structure supports a report generation sequence that mirrors expert writing patterns, as shown in Figure~\ref{fig:dataset_bias}.

To quantitatively assess the impact of organ description order on report quality, we conducted a controlled experiment in which all other prompting variables were held constant, and only the order of organ reasoning steps in the graph was varied. Specifically, we tested all possible permutations of the \blue{n} organs involved in the thought graph traversal graph \blue{(\ie, $n!$ permutations)}. For each permutation, we evaluated the resulting report using six metrics: BLEU-1 through BLEU-4, METEOR, and ROUGE-L.

Let \blue{$\text{ord}_i \in \mathbb{N}^n$} denote the $i$-th permutation of organ order, and let $\text{metric}_{k,i} \in \mathbb{R}_+$ be the corresponding evaluation score for metric $k$ ($k=1,\dots,6$). For each metric $k$, we first identify the index $i_k^\star$ that yields the highest score, defined as  $i_k^\star = \argmax_i \text{metric}_{k,i}$. The permutation corresponding to this index, $\mathbf{b}_k=\text{ord}_{i_k^\star}$, is defined as the best organ order for metric $k$. Then, for each permutation $i$, we compute its Euclidean distance to the best order: $x_{k,i} = \sqrt{\sum_{j=1}^{5} (\text{ord}_{i,j} - \mathbf{b}_{k,j})^2},
$ and plot $x_{k,i}$ (organ order distance) against $\text{metric}_{k,i}$ (performance) to examine the sensitivity of each metric to organ ordering.

We perform both Spearman and Pearson correlation tests to assess statistical dependence. The results show that ROUGE-L is notably sensitive to organ ordering, with a Spearman correlation of $-0.6163$ ($p=8.53 \times 10^{-14}$) and a Pearson correlation of $-0.6072$ ($p=2.45 \times 10^{-13}$), as shown in Figure~\ref{fig:order_metrics_correlation}. This suggests that coherent sequencing of organ descriptions strongly affects linguistic overlap with ground-truth reports and highlights the importance of order modeling in structured report generation.



\blue{\subsection{Failure case analysis}}

\blue{
A common failure mode arises in near-normal or low-abnormality studies. Because TGT explicitly aggregates and refines multiple intermediate reasoning states across traversal steps, weak or spurious cues (for example, subtle projection artifacts or borderline findings) can become repeatedly reinforced during the aggregation process. As a result, the system may assign pathological interpretations to benign variants and generate abnormally specific or overly confident pathological statements in cases that would be considered essentially normal by clinical consensus. In such borderline scenarios, a simpler single-pass prompting baseline often produces a shorter and more conservative report that better aligns with the radiologist’s final assessment.
}

\blue{
From a mechanistic perspective, these failures stem from two interacting factors: (1) the traversal procedure increases the likelihood that minor or ambiguous signals are amplified rather than attenuated, and (2) the model’s tendency to repeatedly elaborate on uncertain cues during iterative querying, which can bias the generated outputs toward pathologically specific interpretations. Although such errors occur in only a small fraction of benchmark cases, they are clinically relevant given the preference for conservative wording in equivocal studies.
}

\blue{
In one representative failure case, the ground-truth report states: ``xxxx sternotomy xxxx and mediastinal postsurgical changes. Stable cardiomegaly. <unk> bronchovascular and interstitial markings xxxx related to low lung volumes and technique. Grossly stable appearance of the lungs compared to prior examination, without overt edema or <unk> airspace consolidation.'' (Here, ``xxxx'' and ``<unk>'' are placeholders introduced by the dataset de-identification process and do not carry clinical meaning.) In contrast, TGT generated a report describing a normal-sized cardiomediastinal silhouette, clear bilateral lung fields, and the presence of pleural effusion in the lower lung zones. This example illustrates a characteristic inconsistency observed in near-normal or postoperative studies: repeated traversal steps can amplify weak or ambiguous cues, leading the model to suppress clinically documented abnormalities (\eg, stable cardiomegaly) while simultaneously introducing findings that are not supported by the visual evidence, such as pleural effusion.
}

\blue{
Several mitigation strategies may help address these failure modes. These include incorporating explicit uncertainty calibration and confidence thresholding to suppress low-confidence refinements; applying post-hoc contradiction checks or introducing a conservative reporting bias (e.g., favoring “no acute cardiopulmonary disease” when confidence is low); limiting the number or depth of traversal steps for borderline cases; and combining TGT outputs with a conservative baseline through ensemble selection or rule-based constraints. In addition, human-in-the-loop review and targeted fine-tuning on near-normal cases may further reduce over-interpretation and improve clinical acceptability.
}

\section{Conclusion}

In this work, we present a test-time scaling framework for chest X-ray report generation that addresses the limitations of shallow reasoning in existing VLLM-based approaches. By introducing a controllable reasoning mechanism and integrating structured medical priors through a thought graph traversal, our method enables stepwise, organ-aware analysis that enhances both the logical consistency and diagnostic correctness of generated reports. Extensive experiments on IU X-Ray and MIMIC-CXR demonstrate that our model significantly outperforms traditional zero-shot and few-shot baselines across multiple evaluation metrics. Ablation studies further confirm the importance of reasoning depth, prompt structure, and organ ordering in improving generation quality. Notably, we show that the order in which organs are analyzed affects model performance, and that optimizing this order can lead to statistically significant gains, especially in metrics like ROUGE-L. \blue{Although our method shows consistent improvements, the evaluation is limited to two English chest X-ray datasets: MIMIC-CXR and IU X-Ray which share similar institutional and linguistic conventions. Reporting styles vary across hospitals, regions, and languages, and broader validation on heterogeneous and multilingual corpora will be necessary to fully establish generalizability. We highlight cross-dataset robustness as an important direction for future work.} Our findings highlight the importance of structured, interpretable reasoning in medical report generation and suggest new directions for controlling inference dynamics in large medical VLLMs.

\section{Acknowledgment}
This work was partially supported by the Key Research and Development Program of Shandong Province under Grant No. 2025CXGC010901.

\bibliographystyle{elsarticle-num}
\bibliography{refs.bib}
\end{document}